\documentclass[preprint]{elsarticle}

\usepackage{multirow}

\begin{document}

\title{On the practice of classification learning for clinical diagnosis and therapy advice in oncology\tnoteref{t1}}
\tnotetext[t1]{\textbf{Declaration of interest:} all three authors are partners of \textit{Autem Medical Research Lab (Brazil)}. We acknowledge support from Autem Medical Research to develop this work.}

\author[ime,autem]{Flavio S Correa da Silva\corref{cor1}}
\ead{flavio.soares@autemmedical.com}

\author[sirio,autem]{Frederico P Costa}
\ead{frederico.costa@autemmedical.com}

\author[esalq,autem]{Antonio F Iemma}
\ead{anfiemma@gmail.com}

\cortext[cor1]{Corresponding author}

\address[ime]{Department of Computer Science, University of Sao Paulo, SP 05508090, Brazil}
\address[sirio]{Oncology Center, Hospital Sirio Libanes - Sao Paulo, SP 01308050}
\address[esalq]{Department of Exact Sciences, University of Sao Paulo - Piracicaba, SP 13418900, Brazil}
\address[autem]{Autem Medical, Bedford NH 03110, USA}

\begin{abstract} %   <- trailing '%' for backward compatibility of .sty file
Artificial intelligence and medicine have a longstanding and proficuous relationship. In the present work we develop a brief assessment of this relationship with specific focus on machine learning, in which we highlight some critical points which may hinder the use of machine learning techniques for clinical diagnosis and therapy advice in practice. We then suggest a conceptual framework to build successful systems to aid clinical diagnosis and therapy advice, grounded on a novel concept we have coined \textit{drifting domains}. We focus on oncology to build our arguments, as this area of medicine furnishes strong evidence for the critical points we take into account here.
\end{abstract}


\maketitle

\textbf{Highlights:}
\begin{itemize}
    \item The relationship between Artificial Intelligence and Medicine is reviewed, highlighting the reasons why few research projects in this field are brought to medical practice.
    \item Focusing on supervised learning applied to diagnosis and therapy plan in oncology, we discuss possible means to overcome this issue.
    \item We present how existing results related to sample complexity in machine learning can be brought to this context to become guidelines to structure and explore results of clinical trials.
    \item We introduce the concept of \textit{drifting domains} as a tool to refine and fine tune the analysis of results obtained with classification learning in our domain of interest.
\end{itemize}

\section{Introduction}

Artificial intelligence and medicine have a longstanding and proficuous relationship, possibly started with the development of the MYCIN system in the early 70s for therapy advice \cite{SHORTLIFFE1973544}.

Medicine has provided the field of artificial intelligence with a plethora of challenging and appealing problems to be solved, particularly in clinical diagnosis (``given a set of signs collected from a patient, select the \textit{best} diagnosis") and in therapy advice (``given an established diagnosis, select the \textit{best} course of actions for treatment"). Artificial intelligence, in turn, has offered promising technologies for problem solving in the medical domain \citep{peek2015thirty}. 

The field of oncology has proven to be particularly fit for modelling and analysis based on artificial intelligence, at least prospectively \citep{kourou2015machine,kaplan2018machine}, due to two major reasons:

\begin{enumerate}
    \item Symptoms in oncology are frequently difficult to identify before later stages of the disease, and cancer can be treated most effectively if identified at early stages of development. Signs of the disease can be diffuse and require high expertise to be selected, collected and analysed. Hence, technologies that can highlight evidence of cancer at early stages are most welcome and challenging at the same time.
    \item Therapy in oncology can be frequently highly aggressive, as it can be based on \textit{drugs} which feature high toxicity and/or \textit{radiotherapy}, which has severe side effects. Hence, technologies that can either refine therapy plans at a personal level to minimise side effects or provide unequivocal evidence of the efficacy of novel therapies are also most welcome.
\end{enumerate}

The practical use of artificial intelligence in medicine, however, has occurred more often in the management of supporting information, rather than in the direct support of the activities of healthcare professionals: medical doctors have been able to augment their capabilities with the support of systems for knowledge representation and processing, as well as automated assistants to process large data chunks aid decision making. However, the practice of automated diagnosis and therapy advice has hardly moved outwith research laboratories to reach everyday activities \citep{chen2017machine}. Some issues can explain why this has happened:

\begin{itemize}
    \item Medicine is strongly regulated by specialised organisations such as the Food and Drug Administration in the US and the European Medicines Agency and CE Mark certifying bodies in the European Economic Area. The levels of detail and transparency required for the description of methods and techniques to be certified by these organisations has proven to be hard, costly and time consuming to achieve, and the effort to reach such levels in the description of novel techniques frequently stays beyond the scope of academic initiatives.
    \item Empirical validation of novel methods and techniques for automated diagnosis and therapy advice requires clinical trials which are highly costly, labour and time consuming. In the medical domain, clinical trials are a required practice for risk mitigation and trust building. Similar procedures are not usual in computer science, and the mismatch in required costs and resources sets apart research initiatives from within the contexts of medical sciences and computer science.
\end{itemize}

In order to bridge the gap between laboratory experimentation and practice in the use of artificial intelligence technologies for clinical diagnosis and therapy advice, these issues must be faced.

In recent years, the field of artificial intelligence has steered towards machine learning, given its significant advances and results \citep{KONONENKO200189}. Following this trend, in the present article we focus on machine learning -- more specifically, supervised classification learning, given that this particularly class of techniques can be formally characterised and analysed in detail -- and how it can be explored to build systems for automated diagnosis and therapy advice which can be used in practice. We consider three specific points related to the issues mentioned in the previous paragraphs:

\begin{enumerate}
    \item One reason for the recent praise to machine learning has been the claim that it decreases subjectivity in artificial intelligence, as domain knowledge and expertise are replaced by statistically grounded data analysis. Critical aspects of domain modelling, however, are still at the core of systems development based on machine learning. Accounting for regulatory issues and transparency requirements, we unveil these aspects and clarify how and why domain knowledge and human expertise -- on the medical domain as well as on computational and statistical techniques -- are central to the design of systems based on machine learning for the medical issues on which we are working.
    \item The demanding requirements of resources connected to clinical trials increase the relevance of controlling sample complexity (in other words, extracting as much information as possible from samples which are kept as small as possible in order to train learning algorithms), which in turn points to the importance of good domain modelling for the problems we are considering. For these reasons, we develop a careful analysis of our population of interest, accounting for the fact that controlled sample cardinality can be key to ensure the feasibility of proposed solutions in practice. 
    \item Our population of interest features an interesting dynamics of evolution, which we believe has not been considered in detail in previous initiatives. We take this dynamics into account and characterise the population as a \textit{drifting domain}, in the sense that it is permanently evolving, although almost always in a smooth rate. We explore how drifting domains can be formally characterised to build accurate predictive models based on classification learning.
\end{enumerate}

The propositions and results presented in this article result from experiments under development at Autem Medical Research, focusing on the development of novel technologies for cancer diagnosis and therapy. In this article we focus on conceptual issues, and refer to our experiments mainly to illustrate our arguments. In a future article we shall present our empirical results from the perspective of advances in clinical diagnosis and therapy advice.

In section \ref{sec:model} we characterise a method and model to build systems based on machine learning for the problems we are considering in medicine, in which we highlight the importance of domain knowledge and expertise for model design. In section \ref{sec:domain} we discuss specific aspects of our domain of interest, including how and why we can focus on populations with finite cardinality and how we can build lower bounds for population cardinality (\textit{aka} sample complexity) given our requirements for precision and reliability. In section \ref{sec:drift} we characterise our proposed notion of \textit{drifting domains} and how it can be used to increase precision and reliability of our results. Finally, in section \ref{sec:conclusion} we present a discussion and conclusions.

\section{Classification learning for diagnosis and therapy advice}
\label{sec:model}

The task of clinical diagnosis can be characterised as a set of steps:

\begin{enumerate}
    \item An individual patient $p$ comes to the medical doctor. The doctor selects a set of signs $S_1$ to observe and analyse. The choice of the set $S_1$ is based on:
    \begin{itemize}
        \item Previous experience and scholarly knowledge.
        \item Tacit selection of a reference population $P$, assuming of course that $p \in P$.
    \end{itemize}
    \item Given the observations related to signs $S_1$, the doctor builds a preliminary set of hypotheses $D$ about the diagnostics of $p$. These hypotheses are dependent upon \textit{previous experience}, \textit{scholarly knowledge} and the \textit{reference population}, which determine an (unknown yet clearly defined) upper bound on the precision and reliability of diagnostics.
    \item Hypotheses are prioritised according to 
    \begin{itemize}
        \item strength of evidence indicating each hypothesis, and
        \item the severity of corresponding diseases.
    \end{itemize}
    Following these priorities, for each hypothesis:
    \begin{enumerate}
        \item A second set of signs $S_2$ is selected. Again, the choice of $S_2$ is based on previous experience, scholarly knowledge and the reference population.
        \item Given the observations related to signs $S_2$, the doctor tags the corresponding hypothesis as either possible or discarded.
    \end{enumerate}
    \item The medical doctor then proceeds to perform information fusion about all hypotheses at hand and final decision about diagnostics.
\end{enumerate}

We have focused on the automation of steps 1 to 3 in clinical diagnoses as outlined in the previous paragraphs. These steps inherently depend upon expert knowledge, regardless of the computational techniques that can be employed for automation.

Required general steps to automate this procedure using supervised classification learning can be characterised as follows:

\begin{enumerate}
    \item Given an individual patient $p$, a reference population $P$ is explicitly characterised, such that $p \in P$.
    \item Based on expert knowledge, a set or preliminary signs $S_1$ is selected.
    \item An oracle $\hat{P} \subseteq P$ containing information about the correlation between signs in $S_1$ and hypotheses in $D$ for diagnostics for population $P$ is retrieved from a database of oracles. An oracle is a collection of pairs $\langle \vec{S}_i, D_i \rangle$ in which $\vec{S}_i \in S_1$ is a tuple of signs and $D \in \{\top,\bot\}$ is a corresponding diagnostics as observed in a patient $p_i \in P$. 
    
    The cardinality of  oracle $\hat{P}$ (i.e. the number of patients $p_i \in \hat{P}$) must be sufficiently large to ensure appropriate levels of precision and reliability of diagnostics performed about any patient $p \in P$, which correspond to a sufficiently large similarity level between empirical classifiers and the best available classifier as determined by unknown upper bounds provided by human expertise.
    \item The correlation between signs and hypotheses is characterised with respect to a family of functions that best captures how decision procedures can be optmised for this correlation. In machine learning and statistics jargon, such families of functions are called kernel functions. The choice of the appropriate family of kernels (e.g. polynomial, gaussian, sigmoid etc. \citep{scholkopf2001learning}) is based on visual inspection of the correlation graphs and expert knowledge about the methods and techniques used to build models for machine learning. The choice of a family of kernels determines another (unknown) upper bound on the precision and reliability of diagnostics.
    \item Hypotheses are prioritised and, for each hypothesis, a second set of signs $S_2$ is selected, based on expert knowledge and the reference population.
    \item A second family of kernels is selected given the observed correlation between $S_2$ and the corresponding hypothesis.
    \item Samples of appropriate cardinality are selected from the population $P$, considering lower bounds for the cardinality of samples as a function of the required precision and reliability of diagnostics that can ensure sufficiently high similarity between empirical classifiers and the best available classifier. These lower bounds can be characterised based on existing theoretical results as detailed in section \ref{sec:domain}. If a sample has cardinality above the identified upper bounds, we have statistical guarantees that obtained classifiers will be, with high probability, sufficiently close to the best available classifier given the upper bounds determined by the expert choices as identified in the previous steps.
    \item Automated decision procedures are built for the diagnosis of patient $p \in P$ employing a sample $\hat{P}$ of appropriate cardinality and the sets of signs $S_1$ and $S_2$.
    \item Decision procedures are employed to build information to support the medical doctor in diagnostics.
\end{enumerate}

Similarly, the task of therapy advice can be characterised as a set of steps:

\begin{enumerate}
    \item An individual patient $p$ comes to the medical doctor, featuring a previously identified most likely diagnostics. The doctor selects a set of tests $T$ to perform, in order to decide for a therapy plan. As in diagnosis, the set $T$ is based on:
    \begin{itemize}
        \item Previous experience and scholarly knowledge.
        \item Tacit selection of a reference population $P$ such that $p \in P$.
    \end{itemize}
    \item Given the outcomes of $T$, the doctor builds a personalised therapy plan for patient $p$. This plan can contain additional decision points in the form of \textit{IF-THEN} rules.
    \item The effectiveness of the treatment is assessed based on empirical observation of attributes which are determined based on expert knowledge and the reference population.
\end{enumerate}

General steps to automate this procedure can be characterised as follows:

\begin{enumerate}
    \item Given a patient $p$ and corresponding most likely diagnostics, and given the reference population $P$ employed in the analysis, alternative therapy plans $T_i$ are ranked according to previous empirical results. Ranking is based on (most likely non-linear) correlation analyses between different therapy plans and their corresponding effective measurements, which are built using samples $\hat{P} \subseteq P$ of appropriate cardinality, which must be such that precision and reliability can be ensured with respect to the best available information about therapy plans. The best available information about different plans, in turn, is based on a history of empirical results. In oncology, given the high mortality related to certain types of tumour, these empirical results can be based on relatively small numbers of cases which are,in turn, described in great detail.
    \item The most highly ranked therapy is selected and applied on $p$.
\end{enumerate}

This characterisation of both clinical diagnosis and therapy advice as sets of steps aims at the identification of the issues that impose limitations on the precision and reliability of diagnosis and therapy advice:

\begin{itemize}
    \item The choice of the sets of signs $S_1$ and $S_2$, as well as the set of therapy plans $T$ depends upon previous experience and scholarly knowledge (\textit{expert knowledge}).
    \item The choice of the reference population $P$ to ground analyses also depends on expert knowledge.
    \item The choice of the family of kernels to characterise the correlations between signs and hypotheses, as well as between diagnostics and therapy plans, depends upon experience and scholarly knowledge about statistical behaviour of specified random variables with respect to stochastic decision procedures in the domain of interest.
\end{itemize}

These limitations are imposed upon the best possible decision procedures for patients in the population $P$. Additionally,

\begin{itemize}
    \item The cardinality of the samples used to build empirical estimates for the best possible decision procedures is determined given lower bounds provided by statistical analysis.
\end{itemize}

Explicit account of these limitations and their reasons are useful to clarify that:

\begin{enumerate}
    \item Most limitations in the quality of automated clinical diagnosis and therapy advice originate from previous experience of medical doctors, scholarly knowledge and the reference population employed to make decisions. These issues are not particular to automated systems and are at place in standard medical practice.
    \item Limitations in precision and reliability of decisions due to sample cardinality can be safely bounded provided that we have access to sufficiently large samples. These bounds, as detailed in the following section, are grounded on scrutinous mathematical analysis.
    
    Moreover,
    \item The dynamics of reference populations can be modelled and taken into account explicitly in decision procedures. In section \ref{sec:drift} we introduce a methodology to provide bounds in precision of decisions given heuristic estimates of the dynamics of the evolution of reference populations.
\end{enumerate}

In the following sections we discuss in detail the lower bounds for the cardinality of samples $\hat{P} \subseteq P$ to estimate decision procedures, as well as the dynamics of the reference populations $P$.

\section{Domain characterisation}
\label{sec:domain}

The initial problem we consider, as characterised in the previous sections, is as follows: given a reference population $P$ of patients featuring sufficiently high homogeneity with respect to the correlation between observable signs $\in S$ and corresponding diagnoses $\in D$, we wish to determine a minimal cardinality $N$ for oracles $\hat{P}$ of the form $\hat{P} = \{\langle\vec{v}_1,d_i\rangle, \ldots, \langle\vec{v}_N,d_N\rangle\}$ such that for the indices $1, \ldots N$ we have that $p_1,  \ldots p_N \in P$, each $\vec{v}_i$ is a tuple of values of signs $\in S$ corresponding to observations about patient $p_i$, and $d_i \in \{\top,\bot\}$ is a confirmed diagnostics for patient $p_i$ with respect to a disease $d \in D$, in which $\top$ indicates \textit{confirmed disease} and $\bot$ indicates \textit{refuted disease}.

We assume a high correlation between values of signs $\vec{v}_i$ and diagnostics $d_i$. We do not assume, however, that this correlation is perfect (which would correspond to a fully deterministic power, at least in a theoretical limit in which information about all patients in $P$ is available, to diagnose disease $d_i$ given observed values of signs $\vec{v}_i$). The set $\{\langle\vec{v}_1,d_1\rangle, \ldots, \langle\vec{v}_{|P|},d_{|P|}\}$ of all pairs $\langle$signs, diagnostics$\rangle \in P$ can, therefore, be \textit{partially inconsistent}, amounting for an (unknown yet determined) upper bound on the precision of diagnoses as well as definition of therapy plans. This upper bound determines the \textit{best available classifiers}. Our task is to build empirical classifiers based on oracles $\hat{P}$ which are provably sufficiently similar to the best available classifiers.

The theory of \textit{Probably Approximately Correct Learning} \citep{valiant1984theory}, as further extended to cope with partial inconsistencies \citep{HAUSSLER199278,mohri2012foundations}, can provide us with lower bounds for the cardinality of $\vec{P}$ as a function of:
\begin{itemize}
    \item $|\vec{\cal V}|$: the cardinality of the valued signs space. Assuming that each sign $s \in S$ can have a finite set of values $\{v_1,\ldots,v_{n_s}\}$ with cardinality $n_s$, we have that $|\vec{\cal V}| = \prod_{s\in S} n_s$.
    \item $\epsilon$: precision, determined as an upper bound for the acceptable disagreement between an empirical classifier and the best available classifier. For example, if $\epsilon = 0.1$, then the probability that, given a tuple of values of signs $\vec{v}$, the empirical classifier and the best available classifier provide the same diagnostics $d \in \{\top,\bot\}$ is at least $90\%$.
    \item $\delta$: reliability, as an upper bound for the risk to build a classifier whose precision parameter is above the specified value $\epsilon$. For example, if $\delta = 0.2$ and $\epsilon = 0.1$, then there is a probability below $20\%$ to select a random classifier built using any oracle $\hat{P}$ with a disagreement below $90\%$ with respect to the best available classifier.
\end{itemize}

Following \cite{mohri2012foundations}, we can define a lower bound for the cardinality of $\hat{P}$ (denoted as $|\hat{P}|$) as:

\begin{center}
    $|\vec{P}| \geq \frac{(ln|\vec{\cal V}|+ln\frac{2}{\delta})}{2\epsilon^2}$
\end{center}

The same lower bound applies for diagnosis and for therapy advice, if we use, for example, \textit{Support Vector Classification} for diagnosis and \textit{Support Vector Correlation} for therapy advice \citep{mohri2012foundations}.

As an example, assuming $|\vec{\cal V}| \approx 150$, we will have $ln|\vec{\cal V}| \approx 5$. This is a realistic assumption, considering the number of parameters and corresponding values which are usually considered by a medical doctor for the tasks under consideration here.

Employing this value, estimates can be obtained for $|\vec{P}|$ given values for $\epsilon$ and $\delta$ as presented in Table \ref{tab:sampleComp}.

\begin{center}
    \begin{table}[ht]
        \centering
        \begin{tabular}{|c|c||c|c|c|}
        \hline
        \multicolumn{2}{|c||}{\multirow{2}{*}{$|\hat{P}|$}} & \multicolumn{3}{|c|}{$\epsilon$} \\
        \cline{3-5}
                    \multicolumn{2}{|c||}{} & 0.1 & 0.2 & 0.3 \\
                    \hline
                    \hline
        \multirow{3}{*}{$\delta$} & 0.1 & 400 & 366 & 346 \\
        \cline{2-5}
        & 0.2 & 100 & 92 & 87 \\
        \cline{2-5}
        & 0.3 & 45 & 41 & 39 \\
        \hline
        \end{tabular}
        \caption{Lower bounds for $|\vec{P}|$ given $\epsilon$ and $\delta$ assuming $|\vec{\cal V}| \approx 150$.}
        \label{tab:sampleComp}
    \end{table}
\end{center}

As a concrete illustration, according to Table \ref{tab:sampleComp}, if we wish to have a probability below $20\%$ that a classifier will be built with disagreement above $10\%$ with respect to the best available classifier, then we need to have access to a reference population with at least $100$ patients.

These results bring existing methods and techniques to the context of clinical diagnosis and therapy advice, and provide medical doctors with concrete parameters and specifications to allow the development of systems based on classification learning for direct support of their activities. They are based, however, on an implicit assumption that any reference population and corresponding oracles for any disease under consideration are static. This assumption is not observed in practice:

\begin{itemize}
    \item Patients pass away, and new patients appear all the time. 
    \item Environmental factors (e.g. pollution rates, dietary habits, stress levels) affect the reference population and the extent to which observed signals correlate with diagnoses.
\end{itemize}

For these reasons, it is reasonable to assume that the reference population undergoes small updates all the time. In order to take into account these updates, we introduce in the next section the concept of \textit{drifting domains} and show how it can be employed to build more refined and precise estimates for clinical diagnosis and therapy advice.

\section{Drifting domains}
\label{sec:drift}

Some implicit assumptions about our domain have been used in the previous sections:

\begin{enumerate}
    \item Our domain is finite, even though it can be large. This way, technical assumptions about probability distributions and logical deductions can be simplified respectively to discrete distributions and propositional reasoning.
    \item Our domain is static and fixed, even though we may not have complete information about each and every element of the domain.
\end{enumerate}

We challenge the second assumption, given our previous consideration that our domain of interest undergoes permanent updates. We assume that these updates are gradual and smooth, as we believe that this assumption is realistic and it simplifies our analyses.

In order to characterise this assumption, we denote domains in which these characteristics are found \textit{drifting domains}. A drifting domain is, therefore: 

\begin{itemize}
    \item A finite domain whose cardinality can be unknown and is permanently updated with small random values which can be positive or negative.
    \item Such that a fixed set of signs characterise each and all elements in the domain.
    \item Such that each sign admits a finite set of values.
    \item Such that the value of each sign associated to each element in the domain is permanently updated, in such way that no ``sudden jump" in a value can be observed.
\end{itemize}

Given these assumptions, and considering that an undetermined time interval occurs between data is collected to build oracles, and that oracles are used for classification of patients, then one reasonable strategy to build more accurate oracles is to assume that, for each value of each sign collected from a patient, the actual present value of that sign can be a different value ``around" the observed value.

One possible way to formalise this strategy is to assume that, for each observed value of a sign, the actual present value is going to be within an interval centred in the observed value. In order to keep calculations simple, we can assume a probability distribution around the observed value, such as e.g. a standardised normal distribution in which the mean value is the observed value. If, additionally, all values are assumed to be discrete approximations of real scalar values, we can further assume that present the values of a sign are, with high probability (above $95\%$), within two standard deviations below and above the observed value.

This way, for each observed value $\tilde{v}$ we build the interval $[\tilde{v}^-,\tilde{v}^+]$ such that $\tilde{v}^- = \tilde{v} - 2\sigma$ and $\tilde{v}^+ = \tilde{v} + 2\sigma$. If we consider the extreme values in this interval, we have for each observed value $\tilde{v}$ the two values $\{\tilde{v}^-,\tilde{v}^+\}$. 

Given $|S|$ as the cardinality of the set of signs, and given one specific observation about one patient belonging to an oracle, reasoning based on extremes of a surrounding interval for the value of each sign builds $2^{|S|}$ alternative ``versions" of that observation. Assuming that the cardinality of a sample is $|\hat{P}|$, we then have a collection of ``possible worlds" $\cal W$ whose cardinality is defined as:

\begin{center}
    $|{\cal W}| = |\hat{P}|2^{|S|}$
\end{center}

If we build one empirical classifier for each possible world, we can test the observations of a new patient with respect to $|\cal W|$ different classifiers. Two possibilities can occur:

\begin{enumerate}
    \item All classifiers agree on the diagnostics for the patient. In this case, this diagnostics is strengthened by being tested considering all variations of the observed sample given the drifting domain under consideration.
    \item We obtain conflicting classifiers across the possible worlds. In this case, upon final decision of the medical doctor, three different strategies can be considered:
    \begin{enumerate}
        \item \textit{Cautious strategy}: the doctor concludes that data is inconsistent and/or insufficient for decision and requires a second cycle of observations, selection of a reference population etc. hoping to be able to resolve the conflict.
        \item \textit{Asymmetric strategy}: in diagnosis \textit{false negatives} can be more harmful than \textit{false positives}, i.e. it can be more damaging to diagnose an unhealthy patient as healthy than the opposite. In this case, if at least one classifier diagnoses the patient as unhealthy, following this strategy the patient can be taken for further examination as potentially unhealthy.
        \item \textit{Uncertainty-based strategy}: some heuristics can be built to assess uncertainty degrees corresponding to the conflicting outcomes of classifiers. For example, some voting procedure can be adopted, such that the confidence on a diagnostics is based on the proportion of classifiers that indicate that diagnostics.
    \end{enumerate}
    
    For diagnosis, any of the three strategies can be adopted. For therapy advice, however, only the first strategy makes sense, given that the selection of the wrong therapy plan is potentially harmful in a symmetric way.
\end{enumerate}

\section{Conclusion}
\label{sec:conclusion}

In this article we have built considerations on how to close the gap between laboratory experimentation and medical practice on using classification learning for clinical diagnosis and therapy advice, with a specific focus on oncology.

More specifically, we have provided an explicit and detailed account of how systems for classification learning can be inserted into the activities workflow of a medical doctor to support diagnosis and therapy advice. Given that an important barrier to the application of machine learning techniques in medicine can be the requirements of large volumes of data, which can point to the necessity of building and running prohibitively costly clinical trials, we have also developed an analysis of sample complexity estimates to build oracles to train systems based on supervised learning, and suggested a pathway to build oracles based on clinical trials of viable dimensions. Finally, we have considered the dynamics of populations from which samples can be taken, and proposed a strategy to refine the analysis of classification results that take into account this dynamics, based on a proposed notion of \textit{drifting domains}.

The considerations we have built here are based on actual experiments under development at Autem Medical Research, where we have worked on novel, lesser aggressive therapies for certain types of cancer and on novel, non-invasive, speedy and low cost technologies for early diagnosis of cancer. Following the guidelines presented here, we have been able to build classifiers to make diagnosis with error rates below $25\%$ based on oracles such that $|\hat{P}| \leq 90$. These classifiers are, at present, undergoing scrutinous analysis and shall be described in a specific article in the near future.

\bibliographystyle{elsarticle-harv}
\bibliography{clinical}

\begin{thebibliography}{10}
\expandafter\ifx\csname natexlab\endcsname\relax\def\natexlab#1{#1}\fi
\expandafter\ifx\csname url\endcsname\relax
  \def\url#1{\texttt{#1}}\fi
\expandafter\ifx\csname urlprefix\endcsname\relax\def\urlprefix{URL }\fi

\bibitem[{Chen and Asch(2017)}]{chen2017machine}
Chen, J.~H., Asch, S.~M., 2017. Machine learning and prediction in medicine:
  beyond the peak of inflated expectations. The New England journal of medicine
  376~(26), 2507.

\bibitem[{Haussler(1992)}]{HAUSSLER199278}
Haussler, D., 1992. Decision theoretic generalizations of the {PAC} model for
  neural net and other learning applications. Information and Computation
  100~(1), 78--150.

\bibitem[{Kaplan et~al.(2018)Kaplan, Berry, Rinn, Ellis, Birchfield, Wahl, Liu,
  Tameishi, Beatty, Dawson, Mehta, Holman, Atwood, Alexander, Bonham, Summers,
  Khalil, Hayete, Wuest, Zheng, Liu, Wang, and Brown}]{kaplan2018machine}
Kaplan, H., Berry, A., Rinn, K., Ellis, E., Birchfield, G., Wahl, T., Liu, X.,
  Tameishi, M., Beatty, J.~D., Dawson, P., Mehta, V., Holman, A., Atwood, M.,
  Alexander, S., Bonham, C., Summers, L., Khalil, I., Hayete, B., Wuest, D.,
  Zheng, W., Liu, Y., Wang, X., Brown, T.~D., 2018. Abstract 5299: Machine
  learning approach to personalized medicine in breast cancer patients:
  Development of data-driven, personalized, causal modeling through
  identification and understanding of optimal treatments for predicting better
  disease outcomes. Cancer Research 78~(13 Supplement), 5299--5299.

\bibitem[{Kononenko(2001)}]{KONONENKO200189}
Kononenko, I., 2001. Machine learning for medical diagnosis: history, state of
  the art and perspective. Artificial Intelligence in Medicine 23~(1), 89--109.

\bibitem[{Kourou et~al.(2015)Kourou, Exarchos, Exarchos, Karamouzis, and
  Fotiadis}]{kourou2015machine}
Kourou, K., Exarchos, T.~P., Exarchos, K.~P., Karamouzis, M.~V., Fotiadis,
  D.~I., 2015. Machine learning applications in cancer prognosis and
  prediction. Computational and structural biotechnology journal 13, 8--17.

\bibitem[{Mohri et~al.(2012)Mohri, Rostamizadeh, and
  Talwalkar}]{mohri2012foundations}
Mohri, M., Rostamizadeh, A., Talwalkar, A., 2012. Foundations of machine
  learning. MIT press.

\bibitem[{Peek et~al.(2015)Peek, Combi, Marin, and Bellazzi}]{peek2015thirty}
Peek, N., Combi, C., Marin, R., Bellazzi, R., 2015. Thirty years of artificial
  intelligence in medicine ({AIME}) conferences: A review of research themes.
  Artificial intelligence in medicine 65~(1), 61--73.

\bibitem[{Scholkopf and Smola(2001)}]{scholkopf2001learning}
Scholkopf, B., Smola, A.~J., 2001. Learning with kernels: support vector
  machines, regularization, optimization, and beyond. MIT press.

\bibitem[{Shortliffe et~al.(1973)Shortliffe, Axline, Buchanan, Merigan, and
  Cohen}]{SHORTLIFFE1973544}
Shortliffe, E.~H., Axline, S.~G., Buchanan, B.~G., Merigan, T.~C., Cohen,
  S.~N., 1973. An artificial intelligence program to advise physicians
  regarding antimicrobial therapy. Computers and Biomedical Research 6~(6),
  544--560.

\bibitem[{Valiant(1984)}]{valiant1984theory}
Valiant, L.~G., 1984. A theory of the learnable. Communications of the ACM
  27~(11), 1134--1142.

\end{thebibliography}

\end{document}